\documentclass{article}
\usepackage{ijcai26}

\usepackage{times}
\usepackage{soul}
\usepackage{url}
\usepackage[hidelinks]{hyperref}
\usepackage[utf8]{inputenc}
\usepackage[small]{caption}
\usepackage{graphicx}
\usepackage{amsmath}
\usepackage{amsthm}
\usepackage{amsfonts}
\usepackage{booktabs}
\usepackage{algorithm}
\usepackage{algorithmic}
\usepackage{tikz}
\usetikzlibrary{positioning,arrows.meta,calc,shapes.geometric,fit,backgrounds}
\usepackage{listings}
\usepackage[switch]{lineno}

\definecolor{codebg}{RGB}{245,245,245}
\definecolor{codeframe}{RGB}{180,180,180}
\lstdefinestyle{code}{
  backgroundcolor=\color{codebg},
  frame=single,
  rulecolor=\color{codeframe},
  basicstyle=\small\ttfamily,
  breaklines=true,
  breakatwhitespace=false,
  keepspaces=true,
  columns=fullflexible,
  xleftmargin=1em,
  xrightmargin=1em,
  framexleftmargin=0.5em,
  framexrightmargin=0.5em,
  aboveskip=1em,
  belowskip=1em,
}
\lstdefinestyle{prompt}{
  backgroundcolor=\color{codebg},
  frame=single,
  rulecolor=\color{codeframe},
  basicstyle=\small\ttfamily,
  breaklines=true,
  breakatwhitespace=false,
  keepspaces=true,
  columns=fullflexible,
  xleftmargin=1em,
  xrightmargin=1em,
  framexleftmargin=0.5em,
  framexrightmargin=0.5em,
  aboveskip=1em,
  belowskip=1em,
}

\newcommand\blfootnote[1]{%
  \begingroup
  \renewcommand\thefootnote{}\footnote{#1}%
  \addtocounter{footnote}{-1}%
  \endgroup
}

\title{Backtrader-Bench: Benchmarking LLM Agents on Algorithmic Trading with Self-Generated MCQs}

\author{
    Ruoxi Zhao
    \and
    Maziar Raissi
    \affiliations
    University of California, Riverside
    \emails
    ruoxi.zhao@email.ucr.edu, maziar.raissi@ucr.edu
}

\begin{document}

\maketitle
\blfootnote{Accepted to the FinLLM Workshop at IJCAI 2026.}

\begin{abstract}
Evaluating LLM coding agents in algorithmic trading is
difficult because static benchmarks risk data contamination and
numerical backtest outputs require ground truth from actual code
execution. We present Backtrader-Bench, a framework with two
complementary pipelines. A deterministic multiple-choice question (MCQ) pipeline generates
questions from backtest configurations across five trading
strategies, 33 templates, and three difficulty tiers, with an
independent checker that re-derives every answer. A
generator-solver filtering pipeline autonomously mines harder
questions: a generator writes questions verified by executable
code, converts them to MCQs, and discards any that a no-tool
solver can answer without code execution. We evaluate 11 models
without tools (10 runs each) and four with-tools configurations
on a 30-question curated set. Tool-augmented agents reach
90.0\% accuracy in a single pass (GPT-5.5 and Opus 4.7),
outperforming the best no-tools baselines (73.0\%, averaged over
10 runs) by 17 percentage points. On 38
separately mined questions, no-tools accuracy drops further,
with half the models falling to roughly random-chance level
(25\%). Beyond evaluation, the scalable MCQ infrastructure is
designed to produce a training corpus for reinforcement learning,
with the ultimate goal of building a specialized agent for
quantitative trading workflows.
\end{abstract}

\section{Introduction}

Large language model (LLM) coding agents equipped with shell
access and iterative code execution can now tackle complex
software-engineering
tasks~\cite{yao:react,schick:toolformer,jimenez:swebench}.
In finance, domain-specific models such as
BloombergGPT~\cite{wu:bloomberggpt} demonstrate strong
financial NLP, reinforcement learning libraries like
FinRL~\cite{liu:finrl} automate strategy optimization, and
tool-augmented systems including
FinAgent~\cite{zhang:finagent},
TradingAgents~\cite{xiao:tradingagents}, and
FinCon~\cite{yu:fincon} show that coupling LLMs with code
execution and multi-agent collaboration can surpass parametric
baselines. Recent surveys chart this rapid progress, covering
financial foundation models broadly~\cite{chen:finfm} and
reasoning models such as DeepSeek-R1 in
finance~\cite{liu:deepseekr1fin}, while noting that rigorous
domain-specific evaluation and reliable tool use remain open
challenges, precisely the gap our benchmark targets. On the evaluation side, benchmarks such as
PIXIU~\cite{xie:pixiu}, FinBen~\cite{xie:finben}, and
InvestorBench~\cite{li:investorbench} cover financial NLP
tasks and decision-making, but none measure whether an agent
can correctly \emph{drive} a backtesting library: run a
backtest under a given configuration, interpret its numerical
output, and report the right answer. This capability matters in practice
because backtesting correctness hinges on subtle interactions
among strategy logic, broker settings, and position-sizing
rules; a single misused parameter produces plausible but wrong
profit-and-loss curves that no amount of surface-level code
fluency can catch. In quantitative finance, such silent errors
carry direct financial consequences: a misstated Sharpe ratio
or understated maximum drawdown can lead portfolio managers to
over-allocate capital to a strategy that would fail under
realistic conditions, turning a backtest artifact into real
trading losses.

Despite a rich ecosystem of Python backtesting frameworks
(backtrader~\cite{backtrader}, zipline~\cite{zipline},
vectorbt~\cite{vectorbt}), existing evaluations target strategy
\emph{generation}.
QuantCode-Bench~\cite{khoroshilov:quantcodebench}, the closest
concurrent work, tests whether an LLM can produce executable
trading code from a natural-language description but does not
test strategy \emph{comprehension}, i.e.\ whether an agent can
run the engine and correctly interpret what it produces.

Building a reliable benchmark faces three obstacles:
\textbf{(1)~Data contamination}: static
benchmarks~\cite{jimenez:swebench,chen:humaneval,austin:mbpp,jain:livecodebench,hendrycks:mmlu,rein:gpqa}
cannot be regenerated once public.
\textbf{(2)~Ground-truth fidelity}: numerical outputs must come
from the same execution path the agent takes.
\textbf{(3)~Quality at scale}: automated generation produces
trivially easy or malformed items, and manual curation does not
scale.

We address all three with \textbf{Backtrader-Bench}, a
contamination-resistant benchmarking framework for
tool-augmented LLM agents on
backtrader~\cite{backtrader}. A \emph{deterministic
multiple-choice question (MCQ) pipeline} runs a backtest, synthesizes stratified
four-option questions across three difficulty tiers and 33
templates spanning five strategies, and re-derives every answer
with an independent checker. Questions are generated at runtime
under a configurable seed, so no answer key need appear on the
public web. A \emph{generator-solver filtering pipeline}
produces harder questions automatically: a generator agent writes
a question with verification code; verified questions become MCQs
with randomized distractors; a no-tool solver discards easy
items; and a tool-augmented solver validates the remainder.

We evaluate 11 models without tools (10 runs each) and four
with-tools agent configurations on a 30-question curated set.
Tool-augmented agents achieve up to 90.0\% accuracy in a single
pass (GPT-5.5 and Opus 4.7), outperforming the best no-tools
baselines (73.0\%, averaged over 10 runs) by 17 percentage
points. On 38 separately mined
questions that the GPT-5.4 filter cannot solve tool-free, no-tools accuracy
drops further, with half the models falling to roughly
random-chance level (25\%). Beyond benchmarking, the ultimate goal of this
scalable MCQ generation infrastructure is to produce a large,
verified training corpus for reinforcement learning, enabling
the development of a domain-specific coding agent for
algorithmic trading.

\paragraph{Contributions.}
\begin{enumerate}
\item A deterministic, self-verifying MCQ pipeline for backtrader
with 33 templates across three difficulty tiers and a per-question
checker that re-runs the backtest.
\item A generator-solver filtering pipeline that autonomously
mines harder questions by discarding those the GPT-5.4 filter
solves tool-free.
\item An empirical evaluation of 11 no-tools models and four
with-tools configurations showing a 17-point accuracy gain from
tool augmentation and near-chance no-tools performance on mined
questions.
\end{enumerate}

\paragraph{Practical relevance.}
The capabilities tested here map directly to quant workflows:
configuring a backtesting engine, running strategy variants, and
interpreting outputs such as drawdown and risk-adjusted returns.
As LLM agents are deployed for strategy prototyping, measuring
whether they can correctly drive a real backtesting library
becomes a prerequisite for safe adoption.

\section{Methodology}
\label{sec:methodology}

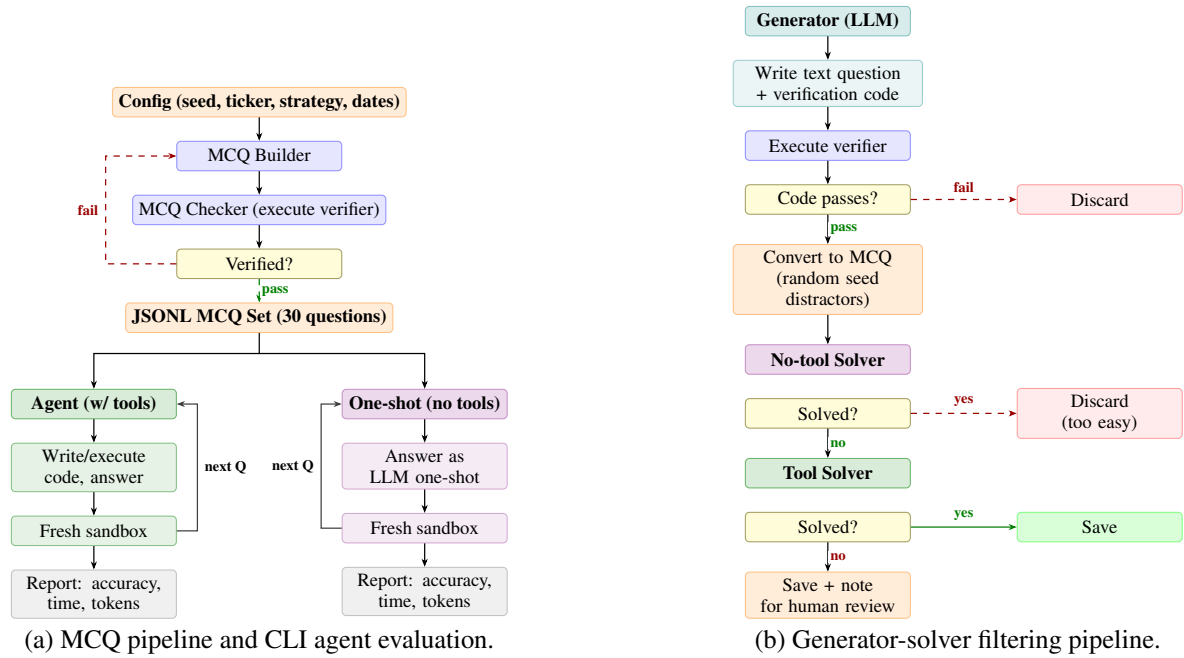
\begin{figure*}[t]
\centering
%% ---- (a) MCQ Pipeline ----
\begin{minipage}[t]{0.48\textwidth}
\centering
\scalebox{0.78}{%
\begin{tikzpicture}[
  node distance=0.4cm,
  every node/.style={font=\footnotesize},
  box/.style={rectangle, draw, rounded corners=2pt,
              minimum height=0.5cm, minimum width=2.8cm,
              text centered, inner sep=3pt},
  arr/.style={-{Stealth[length=4pt]}, semithick},
  failarr/.style={arr, red!60!black, dashed},
  looparr/.style={arr, gray!50!black},
]
\node[box, fill=orange!15, draw=orange!50] (cfg) {\textbf{Config (seed, ticker, strategy, dates)}};
\node[box, fill=blue!10, draw=blue!45, below=of cfg] (build) {MCQ Builder};
\node[box, fill=blue!10, draw=blue!45, below=of build] (checker) {MCQ Checker (execute verifier)};
\node[box, fill=yellow!18, draw=yellow!55!black, below=of checker] (ver) {Verified?};

\draw[arr] (cfg) -- (build);
\draw[arr] (build) -- (checker);
\draw[arr] (checker) -- (ver);
\draw[failarr] (ver.west) -- ++(-1.2,0) |- (build.west)
     node[pos=0.25, left, font=\scriptsize\bfseries, red!60!black] {fail};

\node[box, fill=orange!15, draw=orange!50, below=of ver] (jsonl) {\textbf{JSONL MCQ Set (30 questions)}};
\draw[arr, green!50!black, dashed] (ver) --
     node[midway, right, font=\scriptsize\bfseries,
          green!50!black, fill=white, inner sep=1pt] {pass} (jsonl);

\coordinate (split) at ($(jsonl.south)+(0,-0.35)$);

\node[box, fill=green!50!black!15, draw=green!50!black!60, font=\footnotesize\bfseries,
      below left=0.6cm and 1.4cm of split] (agent) {Agent (w/ tools)};
\node[box, fill=green!50!black!10, draw=green!50!black!50,
      below=of agent, text width=2.4cm, align=center]
     (acode) {Write/execute code, answer};
\node[box, fill=green!50!black!10, draw=green!50!black!50,
      below=of acode] (asand) {Fresh sandbox};
\node[box, fill=gray!12, draw=gray!50,
      below=of asand, text width=2.4cm, align=center]
     (arep) {Report: accuracy, time, tokens};

\draw[arr] (agent) -- (acode);
\draw[arr] (acode) -- (asand);
\draw[looparr] (asand.east) -- ++(0.35,0) |- (agent.east)
     node[pos=0.25, right, font=\scriptsize\bfseries, black] {next Q};
\draw[arr] (asand) -- (arep);

\node[box, fill=violet!15, draw=violet!45, font=\footnotesize\bfseries,
      below right=0.6cm and 1.4cm of split] (oneshot) {One-shot (no tools)};
\node[box, fill=violet!8, draw=violet!35,
      below=of oneshot, text width=2.4cm, align=center]
     (oans) {Answer as LLM one-shot};
\node[box, fill=violet!8, draw=violet!35,
      below=of oans] (bsand) {Fresh sandbox};
\node[box, fill=gray!12, draw=gray!50,
      below=of bsand, text width=2.4cm, align=center]
     (brep) {Report: accuracy, time, tokens};

\draw[arr] (oneshot) -- (oans);
\draw[arr] (oans) -- (bsand);
\draw[looparr] (bsand.west) -- ++(-0.35,0) |- (oneshot.west)
     node[pos=0.25, left, font=\scriptsize\bfseries, black] {next Q};
\draw[arr] (bsand) -- (brep);

\draw[arr] (jsonl.south) -- (split) -| (agent.north);
\draw[arr] (split) -| (oneshot.north);
\end{tikzpicture}%
}

{(a) MCQ pipeline and CLI agent evaluation.}
\end{minipage}%
\hfill
%% ---- (b) Generator-Solver Filtering ----
\begin{minipage}[t]{0.48\textwidth}
\centering
\scalebox{0.78}{%
\begin{tikzpicture}[
  node distance=0.4cm,
  every node/.style={font=\footnotesize},
  box/.style={rectangle, draw, rounded corners=2pt,
              minimum height=0.5cm, minimum width=2.8cm,
              text centered, inner sep=3pt},
  arr/.style={-{Stealth[length=4pt]}, semithick},
  failarr/.style={arr, red!60!black, dashed},
  looparr/.style={arr, gray!50!black},
]
\node[box, fill=teal!15, draw=teal!50, font=\footnotesize\bfseries]
     (gen) {Generator (LLM)};
\node[box, fill=teal!8, draw=teal!40,
      below=of gen, text width=3.0cm, align=center]
     (write) {Write text question + verification code};
\node[box, fill=blue!10, draw=blue!45,
      below=of write] (verify) {Execute verifier};
\node[box, fill=yellow!18, draw=yellow!55!black,
      below=of verify] (vcheck) {Code passes?};

\draw[arr] (gen) -- (write);
\draw[arr] (write) -- (verify);
\draw[arr] (verify) -- (vcheck);

\node[box, fill=red!8, draw=red!40, right=1.8cm of vcheck,
      text width=1.6cm, align=center] (discard1) {Discard};
\draw[failarr] (vcheck.east) -- node[above, font=\scriptsize\bfseries,
     red!60!black] {fail} (discard1);

\node[box, fill=orange!15, draw=orange!50,
      below=0.5cm of vcheck, text width=3.0cm, align=center]
     (mcq) {Convert to MCQ\\(random seed distractors)};

\draw[arr] (vcheck) -- node[midway, right, font=\scriptsize\bfseries,
     green!50!black, fill=white, inner sep=1pt] {pass} (mcq);

\node[box, fill=violet!15, draw=violet!45, font=\footnotesize\bfseries,
      below=0.5cm of mcq] (notool) {No-tool Solver};
\node[box, fill=yellow!18, draw=yellow!55!black,
      below=of notool] (ntcheck) {Solved?};

\draw[arr] (mcq) -- (notool);

\node[box, fill=red!8, draw=red!40, right=1.8cm of ntcheck,
      text width=1.6cm, align=center] (discard2) {Discard\\(too easy)};
\draw[failarr] (ntcheck.east) -- node[above, font=\scriptsize\bfseries,
     red!60!black] {yes} (discard2);

\node[box, fill=green!50!black!15, draw=green!50!black!60, font=\footnotesize\bfseries,
      below=0.5cm of ntcheck] (toolsolv) {Tool Solver};
\node[box, fill=yellow!18, draw=yellow!55!black,
      below=of toolsolv] (tscheck) {Solved?};

\draw[arr] (ntcheck) -- node[midway, right, font=\scriptsize\bfseries,
     green!50!black, fill=white, inner sep=1pt] {no} (toolsolv);

\node[box, fill=green!15, draw=green!50, right=1.8cm of tscheck,
      text width=1.6cm, align=center] (keep) {Save};
\draw[arr, green!50!black] (tscheck.east) -- node[above, font=\scriptsize\bfseries,
     green!50!black] {yes} (keep);

\node[box, fill=orange!15, draw=orange!50, below=0.5cm of tscheck,
      text width=2.4cm, align=center] (human) {Save + note\\for human review};

\draw[arr] (tscheck) -- node[midway, right, font=\scriptsize\bfseries,
     red!60!black, fill=white, inner sep=1pt] {no} (human);

\end{tikzpicture}%
}

{(b) Generator-solver filtering pipeline.}
\end{minipage}

\caption{System workflows. (a) The MCQ pipeline generates and verifies
  questions, then dispatches them to an agent (w/ tools) or one-shot
  (no tools) evaluator. (b) The generator-solver filtering pipeline:
  the generator writes a free-form text question with verification
  code; failed verification discards the question. Verified questions
  are converted to MCQs with randomized distractors. A no-tool solver
  filters out questions solvable without code execution. A
  tool-augmented solver validates the remainder; unsolved questions
  are saved with a note for human review.}
\label{fig:workflows}
\end{figure*}

This section describes the deterministic MCQ pipeline that
generates verified questions from backtrader backtests
(Section~\ref{sec:mcq_pipeline}) and the generator-solver
filtering pipeline that automatically curates question quality
(Section~\ref{sec:filtering_pipeline}).
Figure~\ref{fig:workflows} illustrates both workflows.

\subsection{Data collection: Backtrader\_MCQ pipeline}
\label{sec:mcq_pipeline}

\subsubsection{Trading strategies and indicators}

The MCQ pipeline supports five canonical trading strategies: SMA
Crossover, Rolling Window Mean, EWMA, RSI, and MACD Crossover.
These exercise the core backtrader API (indicators, crossover
signals, position sizing, broker configuration) while keeping
strategy logic transparent enough for unambiguous verification;
more complex strategies are planned as future extensions
(Section~\ref{sec:future}). SMA Crossover is adapted from
backtrader's example code; the remaining four from Pik and
Ghosh~\cite{pik-ghosh:backtesting}. All strategies inherit from
a shared instrumentation base class that logs orders, trades,
and portfolio snapshots (Appendix~\ref{app:strategies}).

\subsubsection{Difficulty taxonomy}

The 33 question templates are each assigned to one of three
difficulty tiers based on the computational steps required:
Easy (9 templates): direct single-value lookups;
Medium (11): conditional filtering or simple
aggregation; Hard (13): multi-step derived metrics
or comparative backtests with alternative parameters.
Templates were designed from backtrader's plotting outputs
(Figure~\ref{fig:backtrader_plot}) and standard quantitative
finance metrics, then implemented with the assistance of
Cursor, an LLM-powered coding agent.
Full tier definitions and example templates are in
Appendix~\ref{app:difficulty}; the complete template list is
in Appendix~\ref{app:templates}.

\subsubsection{Build-checker pipeline}

The pipeline produces verified questions through two stages
(Appendix~\ref{app:pipeline}): a build stage that runs
a Backtrader backtest, generates MCQs from the selected templates
with randomized distractors, and writes JSONL records; and a
check stage that independently re-runs the backtest and
re-derives each answer, flagging any mismatch. Both stages share
the same deterministic engine, so agreement provides a strong
guarantee that the ground truth is internally consistent. This
dual-verification design reflects the standard practice in
quantitative finance, where backtest results are independently
reproduced before any capital allocation decision.

\subsubsection{Sampling and reproducibility}

A single random seed controls all stochastic aspects of question
generation (template selection, distractor generation, option
shuffling, date/parameter sampling); the backtest itself is
deterministic. The 160-question dataset is generated in
``all-templates'' mode (Appendix~\ref{app:config}), covering all
33 templates across three difficulty levels for five strategies.
Fresh question sets can be produced by changing the ticker, date
range, or seed (Appendix~\ref{app:sampling}).

\subsection{Generator-solver filtering pipeline}
\label{sec:filtering_pipeline}

\subsubsection{Motivation}

The deterministic pipeline's templates are hand-designed and
finite. To scale, we let an LLM generate questions autonomously
and filter for quality using two solver agents with complementary
capabilities (Figure~\ref{fig:workflows}b;
Appendix~\ref{app:mining_pipeline}).

\subsubsection{Generator and verification}

A generator agent writes a free-form question together with
self-contained Python verification code that runs a Backtrader
backtest and prints the ground-truth answer
(Appendix~\ref{app:generator_prompt}). The verification code
accepts an environment variable (MCQ\_SEED) that parameterises
inputs such as ticker, date range, and commission rate, enabling
seed-controlled distractor generation. The pipeline executes this
code in a subprocess with a 120-second timeout; failures discard
the question. The generator prompt includes a history of prior
attempts and their outcomes, allowing the LLM to learn from its
own mistakes within a mining session.

\subsubsection{MCQ conversion}

Verified questions are converted into four-option MCQs. Three
distractors are generated by re-running the verification code
with perturbed seed values, collecting unique answer values; if
fewer than four emerge, multiplicative jitter ($\pm$10\%,
$\pm$20\%) fills the gap. Option ordering is shuffled
deterministically. A well-posedness check validates that all
four options are non-empty and the ground-truth answer maps to
exactly one of A/B/C/D.

\paragraph{Distractor calibration.}
Because distractors are formed by multiplicative perturbation of
the correct value (scale factors from $0.7$ to $1.3$ with small
random jitter), every distractor lies within the same order of
magnitude as the correct answer. Across $20{,}000$ simulated
items, distractor values fall between $7\%$ and $33\%$ from the
correct value (mean $20\%$), and the nearest distractor averages
$12.8\%$ away (median $10.7\%$). Coarse-scale or
order-of-magnitude elimination is therefore ineffective:
selecting the correct option requires computing the precise
numerical value rather than estimating its rough size.

\subsubsection{Multi-stage filtering}

\paragraph{Stage 1: No-tool solver (discard if solved).}
Each MCQ is presented to a no-tool solver (Cursor SDK with all
tool use disabled and a runtime inspector that cancels on any
tool invocation). If the solver answers correctly, the question
is discarded as solvable from parametric knowledge alone.

\paragraph{Stage 2: Tool-augmented solver (keep if solved).}
Surviving questions are passed to a tool-augmented solver with
full code-execution permissions. If it answers correctly, the
question is kept: the GPT-5.4 filter cannot answer it tool-free,
and it is confirmed well-posed.

\paragraph{Stage 3: Human review.}
Questions that neither solver answers are saved for human review
to determine whether they are genuinely difficult or malformed.

\subsubsection{Agent backends and persistence}

Both solver roles and the generator are instantiated through a
uniform adapter layer supporting four coding-agent CLIs (Cursor,
Claude Code, OpenAI Codex, GitHub Copilot); all experiments use
the Cursor backend (Appendix~\ref{app:agent_backends}). Pipeline
state is persisted as flat JSONL files with SHA-256 content-based
deduplication across mining rounds and well-posedness validation
before filtering (Appendix~\ref{app:mining_pipeline}).

\section{Evaluation Framework}

\subsection{Prompting design}

Both evaluation modes receive the same MCQ prompt, which includes
the backtest configuration (ticker, date range, strategy,
parameters, commission rate, initial cash), the question text,
four labeled options (A/B/C/D), and an answer format instruction
requesting the response as a single letter (A, B, C, or D).
Example questions
for each difficulty tier are shown in
Appendix~\ref{app:examples}.

\paragraph{With-tools mode.}
The agent is invoked via the Cursor Agent CLI with full
code-execution permissions. The prompt instructs the agent to
write a Python solution script, execute it in an isolated
sandbox, and reply with the answer letter. The agent may install
packages, read/write files, and iterate on its solution across
multiple tool calls. Each question runs in a fresh working
directory to prevent information leakage between questions
(Appendix~\ref{app:invocation}; prompt in
Appendix~\ref{app:prompts}).

\paragraph{No-tools mode.}
The agent is invoked via the Cursor SDK with all tool use
disabled through three enforcement layers: (1) SDK-level
configuration that removes all MCP servers, subagents, and
ambient settings; (2) a prompt instruction explicitly
prohibiting code execution and all external tools; and (3) a
runtime stream inspector that monitors every event and
immediately cancels the run if any tool invocation is detected.
The agent must answer from parametric knowledge alone
(Appendix~\ref{app:invocation}).

\subsection{Models}

\paragraph{With-tools models.}
We evaluate four model configurations available through the
Cursor Agent CLI: GPT-5.5, Opus 4.7, auto (default routing),
and composer-2. Each model runs in a sandbox with identical
tool permissions and timeout settings.

\paragraph{No-tools models.}
We evaluate 11 models spanning five providers: Opus 4.7, Gemini
3.1 Pro, GPT-5.3 Codex, GPT-5.5, Sonnet 4.6, Opus 4.6,
Grok 4.3, Kimi K2.5, auto (default), composer-2, and Gemini 2.5
Flash. All models are accessed through the Cursor SDK using the
same no-tools enforcement mechanism. Each model is run 10 times
on the curated benchmark to account for LLM stochasticity;
reported accuracies are 10-run averages.

\subsection{Datasets}

\paragraph{Curated benchmark (30 questions).}
A balanced subset of 30 questions (10 easy, 10 medium, 10 hard)
drawn from the 160-question base pool generated by the
deterministic MCQ pipeline. All questions use AAPL data from
2020-01-01 to 2024-01-01 with \$50,000 initial cash and 0.1\%
commission. This set is used for both with-tools and no-tools
evaluation.

\paragraph{Mined question set (38 questions).}
Produced by 100 mining attempts (generator: GPT-5.5-medium;
solvers: GPT-5.4). Of 98 well-posed questions, 60 were discarded
as too easy and 38 accepted. This set
is used for no-tools quality validation: all 11 models are run
once to measure how much filtering reduces accuracy.

\subsection{Metrics}

We report accuracy (\%) overall and per difficulty tier, total elapsed
time, and standard deviation across 10 runs (no-tools curated
benchmark only). Full protocol details are in
Appendix~\ref{app:eval_protocol}.

\section{Results}

\subsection{Curated benchmark results}

Tables~\ref{tab:with-tools} and~\ref{tab:no-tools} present
accuracy on the curated 30-question benchmark (10 easy, 10
medium, 10 hard). No-tools results are averaged over 10
independent runs per model; with-tools entries reflect a single
evaluation pass.

\begin{table}[h!]
\centering
\small
\caption{With-tools accuracy (\%). Each agent writes, executes,
  and debugs code per question in an isolated sandbox.}
\label{tab:with-tools}
\begin{tabular}{lccccc}
\toprule
Model & Easy & Med. & Hard & All & Min. \\
\midrule
GPT-5.5        & 100 & 100 & 70 & \textbf{90.0} & \textbf{83} \\
Opus 4.7       & 100 & 100 & 70 & \textbf{90.0} & 103 \\
auto (default) & 100 &  90 & 60 & 83.3 & \textbf{83} \\
composer-2     & 100 &  80 & 50 & 76.7 & 85 \\
\bottomrule
\end{tabular}

\vspace{1em}

\caption{No-tools accuracy (\%), averaged over 10 runs per model.
  Each agent answers in a single API call with no code execution.}
\label{tab:no-tools}
\begin{tabular}{lccccc}
\toprule
Model & Easy & Med. & Hard & All & Min. \\
\midrule
Opus 4.7       & 100 & 51 & 68 & \textbf{73.0} & \textbf{1.1} \\
Gemini 3.1 Pro &  97 & 68 & 54 & \textbf{73.0} & 6.3 \\
GPT-5.3 Codex  &  99 & 84 & 25 & 69.3 & 2.0 \\
GPT-5.5        & 100 & 68 & 36 & 68.0 & 8.0 \\
Sonnet 4.6     & 100 & 58 & 44 & 67.3 & 2.8 \\
Opus 4.6       & 100 & 50 & 43 & 64.3 & 6.7 \\
Grok 4.3       &  87 & 54 & 37 & 59.3 & 3.4 \\
Kimi K2.5      &  75 & 56 & 36 & 55.7 & 1.8 \\
auto (default) &  76 & 63 & 20 & 53.0 & 2.2 \\
composer-2     &  72 & 61 & 20 & 51.0 & 2.8 \\
Gemini 2.5 Flash &  68 & 46 & 33 & 49.0 & 1.8 \\
\bottomrule
\end{tabular}
\end{table}

\paragraph{Tool augmentation improves accuracy.}
The best with-tools configurations (GPT-5.5 and Opus 4.7, both
90.0\% in a single pass) outperform the best no-tools results
(Opus 4.7 and Gemini 3.1 Pro, both 73.0\% over 10 runs) by 17
percentage points. All four
with-tools models achieve 100\% on easy questions. The gap
widens on harder tiers: with-tools agents score 50\% to 70\% on
hard questions, while no-tools models range from 20\% to 68\%.
For practitioners, this means that an agent without code
execution will misreport metrics like drawdown or trade count on
roughly one in four questions, errors that could propagate into
flawed strategy selection if left unchecked.

\paragraph{Difficulty tiers discriminate as intended.}
Across most models, accuracy decreases monotonically from easy
to medium to hard, mirroring the progression from simple lookups
(e.g., closing price on a date) through conditional queries
(e.g., first profitable trade) to multi-step derived metrics
(e.g., maximum drawdown, comparative Sharpe ratio). A notable
exception is Opus 4.7 (no-tools), which scores higher on hard
(68\%) than medium (51\%) over 10 runs, suggesting its extended
reasoning helps more on complex numerical questions than on
intermediate ones.

\paragraph{Model rankings shift between modes.}
Opus 4.7 ties for first in both modes (90.0\% with tools,
73.0\% without). However, rankings are not preserved: GPT-5.3
Codex ranks third without tools (69.3\%) but is not among the
with-tools models tested. Composer-2 ranks last in both modes,
confirming that weaker models benefit less from tool access.

\paragraph{Multi-run averaging matters.}
No-tools standard deviations range from 1.1 to 8.0 percentage
points, confirming that single-pass evaluation can be misleading.
Opus 4.6 scored 76.7\% in a single run but averages 64.3\% over
10 runs, an overestimate of more than 12 points.

\paragraph{Cost of tool use.}
With-tools runs require 83-103 minutes per 30-question set;
no-tools runs complete in 1-8 minutes (10-90$\times$ faster).

\subsection{Mining pipeline results}

Of 100 mining attempts, 98 produce well-posed questions; the
pipeline accepts 38 (39\%) where the no-tools solver fails but
the tool solver succeeds, and discards 60 as too easy. The
questions span eight domains
(Appendix~\ref{app:mcq-domains}). Bracket/stop-order questions
have the highest acceptance rate (9/15, 60\%) because they depend
on backtrader's event-loop and order-fill mechanics. MA crossover
questions are most numerous (43) but only 47\% are accepted, as
the no-tools solver often infers the answer. Weaker domains such
as slippage (1/6) and analyzer metrics (1/5) remain answerable
without code execution. These findings have practical
implications for quant teams: the questions that most reliably
resist tool-free solving are precisely the order-mechanics scenarios
where errors carry the highest financial risk, such as
misconfigured stop-losses or unexpected order cancellations.

\subsection{Mined MCQ quality validation}

Table~\ref{tab:mined-notools} evaluates all 11 no-tools models
on the 38 accepted mined MCQs. Compared to the curated benchmark
(Table~\ref{tab:no-tools}), no-tools accuracy drops across every
model, confirming that the filtering pipeline produces genuinely
harder questions.

\begin{table}[h!]
\centering
\small
\caption{No-tools accuracy (\%) on the 38 accepted mined MCQs.
  Random-chance baseline is 25\%.}
\label{tab:mined-notools}
\begin{tabular}{lcc}
\toprule
Model & Acc.\,\% & Time\,(min) \\
\midrule
Gemini 3.1 Pro   & \textbf{60.5} &  9.4 \\
GPT-5.5          & \textbf{60.5} & 10.7 \\
Opus 4.6         & 57.9 &  5.7 \\
Opus 4.7         & 52.6 & \textbf{2.3} \\
Sonnet 4.6       & 39.5 &  4.8 \\
GPT-5.3 Codex    & 34.2 &  9.7 \\
Gemini 2.5 Flash & 29.0 &  2.5 \\
auto (default)   & 26.3 & 16.0 \\
composer-2       & 23.7 &  4.4 \\
Grok 4.3         & 21.1 &  5.6 \\
Kimi K2.5        & 15.8 &  3.8 \\
\bottomrule
\end{tabular}
\end{table}

\paragraph{Accuracy drops substantially across all models.}
Every model performs worse on the mined set than on the curated
benchmark, confirming that the filtering pipeline successfully
removes questions answerable from parametric knowledge. The
largest drops are for mid-tier models: Grok 4.3 falls from
59.3\% to 21.1\% (38 points), Kimi K2.5 from 55.7\% to 15.8\%
(40 points), and composer-2 from 51.0\% to 23.7\% (27 points).
Even the strongest models decline: Gemini 3.1 Pro drops from
73.0\% to 60.5\%, and GPT-5.5 from 68.0\% to 60.5\%.
The results reveal a clear split: the top four models (Gemini 3.1
Pro, GPT-5.5, Opus 4.6, Opus 4.7) cluster between 53\% and 61\%,
while five models (Gemini 2.5 Flash, auto (default), composer-2,
Grok 4.3, Kimi K2.5) fall to between 16\% and 29\%. Three of
these land near the 25\% random-chance baseline; at $n=38$ their
scores are statistically indistinguishable from chance, so
without code execution they are effectively guessing. This
validates the core premise of the mining pipeline: tool-free
reasoning collapses toward chance for most models and plateaus
near 60\% even for the strongest, far below tool-augmented
performance and bounded by the strength of the filter model.

\paragraph{Filter model determines difficulty ceiling.}
The no-tools filter used during mining was GPT-5.4. Stronger
models such as Gemini 3.1 Pro and GPT-5.5 still score 60.5\%
on these supposedly tool-requiring questions, indicating that
mined-question difficulty is relative to the filtering model.
Using a stronger filter or combining curated MCQ templates with
the generator via in-context learning
(Section~\ref{sec:future}) should improve difficulty calibration.

\paragraph{Error analysis: which financial concepts are hardest.}
Agents most consistently err on multi-step derived financial
metrics: maximum drawdown (tracking a running peak across the
equity curve), exposure-adjusted returns (computing time-in-market
from position logs), and risk-adjusted ratios such as Sharpe and
Sortino. On the mined set, bracket-order and trailing-stop
questions produce the largest accuracy drops because they depend
on backtrader's stateful event-loop and order-fill mechanics. These
patterns suggest that failures concentrate in multi-step numerical
computation under library-specific execution semantics, not in
concept recognition.

\section{Limitations}
\label{sec:limitations}

\paragraph{Filter model and evaluation variance.}
Mined-question difficulty is relative to the filtering model
(GPT-5.4 in our experiments), not absolute;
stronger models still solve 60.5\% without tools. Both mined MCQ
results (single-run) and with-tools results (single-pass due to
83-103 min per set) would benefit from multi-run averaging.

\paragraph{Computational cost.}
The mining pipeline is token-intensive: each attempt requires the
generator to produce a question and verification code, the
verifier to execute it, and both solvers to write and run their
own code. A single 100-attempt mining session consumes on the
order of millions of tokens across generator and solver roles and
takes several hours of total elapsed time. With-tools evaluation is
similarly expensive (83-103 minutes per 30-question set), as the
agent iteratively writes, executes, and debugs code for each
question. These costs limit the scale of both mining and
evaluation: no-tools results on the 38 mined MCQs are single-pass
rather than 10-run averages, and we did not run with-tools
evaluation on the mined set at all.

\paragraph{Sample size and statistical power.}
The limited scale of our evaluation (30 curated and 38 mined
questions) is a direct consequence of the per-run compute cost
noted above. At $n=30$, Wilson $95\%$ confidence intervals are
wide (roughly $\pm 15$ points), so small differences among the
top no-tools models (73.0\%, 69.3\%, and 68.0\%) are not
statistically significant (Fisher exact $p > 0.75$), and we do
not interpret fine-grained rankings. The effects we emphasize are
those that survive this uncertainty: the large tool-free
degradation on the mined set (top models fall to 60.5\%, weaker
models to roughly chance) and its consistency across all four
paired tool-augmented configurations and 10 no-tools runs.

\paragraph{Coverage and generalizability.}
The curated pipeline covers five canonical strategies and 33
templates; the mining pipeline adds eight domains but 44\% are MA
crossover variants. All experiments use Cursor as the sole agent
framework and backtrader as the sole library, so we cannot fully
isolate model capability from framework or library familiarity.
The framework operates on historical single-equity data (AAPL,
2020-2024) and does not test live execution or multi-asset
portfolios.

\section{Conclusion and Future Work}

We presented Backtrader-Bench, a benchmarking framework with two
complementary pipelines: a deterministic MCQ pipeline (33
templates, three difficulty tiers) and a generator-solver
filtering pipeline that mines harder questions by discarding
those the GPT-5.4 filter solves tool-free.

Tool augmentation improves accuracy by 17 points on the curated
set (single-pass with-tools 90.0\% vs.\ 10-run no-tools 73.0\%). On mined questions, half the no-tools
models fall to roughly chance level, validating the filtering
pipeline. Mined-question difficulty is relative to the filter
model, indicating room for improvement. These pipelines serve as
infrastructure for building a domain-specific trading agent via
reinforcement learning on the structured feedback each MCQ
provides.

\subsubsection{Future research}
\label{sec:future}

\paragraph{Improving mining yield and difficulty.}
Feeding the generator in-context examples of accepted (hard) and
discarded (easy) questions could steer it toward domains with
high acceptance rates such as bracket-order mechanics (60\%).
Replacing the no-tools filter (GPT-5.4) with a stronger model
or ensemble would raise the difficulty ceiling.

\paragraph{Scaling coverage.}
Adding strategies (Bollinger Bands, pairs trading, momentum),
hard-tier metrics (Sharpe, Sortino, Calmar), additional
backtesting frameworks (zipline~\cite{zipline},
vectorbt~\cite{vectorbt}, QSTrader~\cite{qstrader}), and
cross-agent benchmarking (Claude Code, Codex, Copilot) would
broaden coverage and disentangle library/framework confounds.
Scaling the curated set toward roughly 100 balanced questions
(matched across easy, medium, and hard tiers) spanning additional
tickers, date ranges, and strategy parameters, together with
multi-run averaging on the mined and with-tools settings, would
tighten the confidence intervals reported in
Section~\ref{sec:limitations} and improve the stability of
per-model comparisons.

\paragraph{Component ablations.}
Controlled ablations would quantify each pipeline component's
contribution: removing the independent backtest checker, replacing
generator-solver filtering with direct template-based hard
questions, and disabling SHA-256 deduplication. Measuring the
resulting change in question quality and discriminative power
under each condition would empirically validate the design choices
this paper currently motivates qualitatively.

\paragraph{RL fine-tuning of a domain-specific agent.}
The ultimate goal is to build a domain-specific coding agent for
algorithmic trading. Each MCQ carries structured feedback
(correct/incorrect, difficulty tier, domain, tool use required)
that can serve as a dense reward signal. As the mining pipeline
scales to thousands of verified questions, it could produce
enough training data to fine-tune an agent that reliably drives
backtest engines and avoids the configuration errors that current
general-purpose models frequently make. For quantitative finance
teams, such a specialized agent would reduce the manual effort
of validating backtest outputs and lower the risk of silent
errors propagating into capital allocation decisions.

\section{Data and Code Availability}

The 160-question dataset, both pipelines, evaluation scripts, and
configuration files are publicly available at \url{https://github.com/rzhao999/Backtrader-Bench}, with
instructions to regenerate datasets under new seeds and reproduce
all experiments.

\section*{Ethical Statement}

The framework benchmarks agents against historical price data and does not take live positions; it is intended as a measurement tool, not an automated trading system. Agents may produce trading-shaped output during evaluation. Users who repurpose the strategy code
beyond benchmarking should follow standard risk-management and regulatory practice; the authors accept no responsibility for financial losses incurred by such use.

%% References
\bibliographystyle{named}
\bibliography{backtrader_bib}

\newpage
\onecolumn
\appendix
%\leftlinenumbers

\section{Strategy Implementations}
\label{app:strategies}

The following listings show the core logic of each trading strategy
used in the MCQ pipeline. All strategies inherit from a shared
instrumentation base class that logs order executions, trade
closures, and daily portfolio snapshots. Each strategy is built
around one or more standard technical indicators, described below.

\paragraph{SMA Crossover.}
The \textbf{Simple Moving Average} (SMA) is the unweighted
arithmetic mean of the last $n$ closing prices, where $C_t$
denotes the closing price at time $t$:
\begin{linenomath*}
\[
  \text{SMA}_t(n) \;=\; \frac{1}{n}\sum_{i=0}^{n-1} C_{t-i}
\]
\end{linenomath*}
This strategy computes two SMAs with different lookback
periods: a fast (short-term, default 10 days) and a slow
(long-term, default 30 days). A buy signal fires when the fast
SMA crosses above the slow SMA, indicating upward momentum; a
sell signal fires on the reverse crossing.
{\small\begin{verbatim}
class SmaCross(McqInstrumentedStrategy):
    params = dict(pfast=10, pslow=30, target_percent=1.0)

    def __init__(self):
        self.sma_fast = bt.ind.SMA(self.data.close, period=self.p.pfast)
        self.sma_slow = bt.ind.SMA(self.data.close, period=self.p.pslow)
        self.crossover = bt.ind.CrossOver(self.sma_fast, self.sma_slow)

    def next(self):
        if not self.position and self.crossover > 0:
            self.order = self.buy()
        elif self.position and self.crossover < 0:
            self.order = self.close()
\end{verbatim}}

\paragraph{Rolling Window Mean.}
Uses a single SMA with a longer lookback window (default 90 days)
as a trend filter. The rolling mean smooths out short-term noise,
providing a baseline for the recent price regime. The strategy
buys when the closing price rises above the rolling mean
(interpreted as an uptrend) and exits when it falls below
(interpreted as a downtrend).
{\small\begin{verbatim}
class RollingWindowMean(McqInstrumentedStrategy):
    params = dict(rolling_window=90, target_percent=1.0)

    def __init__(self):
        self.rolling_mean = bt.ind.SMA(self.data.close,
                                       period=self.p.rolling_window)

    def next(self):
        if self.data.close[0] > self.rolling_mean[0] and not self.position:
            self.buy()
        elif self.data.close[0] < self.rolling_mean[0] and self.position:
            self.close()
\end{verbatim}}

\paragraph{EWMA.}
The \textbf{Exponential Moving Average} (EMA) assigns
exponentially decreasing weights to older prices, making it more
responsive to recent data than the SMA. With smoothing factor
$\alpha = 2/(n+1)$, the EMA is defined recursively:
\begin{linenomath*}
\[
  \text{EMA}_t \;=\; \alpha\, C_t \;+\; (1 - \alpha)\,\text{EMA}_{t-1}
\]
\end{linenomath*}
This strategy uses a fast EMA (default period 5) and a slow EMA
(default period 30). It buys when the fast EMA rises above the
slow EMA and sells on the reverse crossing.
{\small\begin{verbatim}
class ExponentialWeightedMovingAverage(McqInstrumentedStrategy):
    params = dict(pfast=5, pslow=30, target_percent=1.0)

    def __init__(self):
        self.ema_fast = bt.ind.EMA(self.data.close, period=self.p.pfast)
        self.ema_slow = bt.ind.EMA(self.data.close, period=self.p.pslow)

    def next(self):
        if self.ema_fast[0] > self.ema_slow[0] and not self.position:
            self.buy()
        elif self.ema_fast[0] < self.ema_slow[0] and self.position:
            self.close()
\end{verbatim}}

\paragraph{RSI.}
The \textbf{Relative Strength Index} (RSI) is a momentum
oscillator that measures the speed and magnitude of recent price
changes on a 0 to 100 scale:
\begin{linenomath*}
\[
  RS \;=\; \frac{\overline{U}_n}{\overline{D}_n}\,,
  \qquad
  \text{RSI} \;=\; 100 - \frac{100}{1 + RS}
\]
\end{linenomath*}
where $\overline{U}_n$ is the average gain and $\overline{D}_n$
the average loss over the last $n$ periods. Low RSI values
indicate oversold conditions; high values indicate overbought
conditions. This strategy buys when RSI drops below 10 (deeply
oversold) and sells when it rises above 90 (deeply overbought).
Default lookback period: 12 days.
{\small\begin{verbatim}
class RsiStrategy(McqInstrumentedStrategy):
    params = dict(rsi_period=12, rsi_buy_threshold=10,
                  rsi_sell_threshold=90, target_percent=1.0)

    def __init__(self):
        self.rsi = bt.ind.RSI_Safe(self.data.close, period=self.p.rsi_period)

    def next(self):
        if self.rsi[0] < self.p.rsi_buy_threshold and not self.position:
            self.buy()
        elif self.rsi[0] > self.p.rsi_sell_threshold and self.position:
            self.close()
\end{verbatim}}

\paragraph{MACD Crossover.}
The \textbf{Moving Average Convergence Divergence} (MACD)
indicator tracks the difference between a fast EMA and a slow EMA
of the closing price, with a separate signal line that is itself
an EMA of the MACD:
\begin{linenomath*}
\[
  \text{MACD}_t \;=\; \text{EMA}_{\text{fast}}(t)
                 \;-\; \text{EMA}_{\text{slow}}(t)\,,
  \qquad
  \text{Signal}_t \;=\; \text{EMA}_{s}\!\bigl(\text{MACD}_t\bigr)
\]
\end{linenomath*}
When the MACD line crosses above the signal line, it suggests
bullish momentum; a downward crossing suggests bearish momentum.
Default periods: 12 (fast EMA), 26 (slow EMA), 9 (signal EMA).
{\small\begin{verbatim}
class MacdCrossoverStrategy(McqInstrumentedStrategy):
    params = dict(pfast=12, pslow=26, psignal=9, target_percent=1.0)

    def __init__(self):
        self.macd = bt.ind.MACD(self.data.close,
                                period_me1=self.p.pfast,
                                period_me2=self.p.pslow,
                                period_signal=self.p.psignal)
        self.macd_cross = bt.ind.CrossOver(self.macd.macd, self.macd.signal)

    def next(self):
        if self.macd_cross > 0 and not self.position:
            self.buy()
        elif self.macd_cross < 0 and self.position:
            self.close()
\end{verbatim}}

\section{Difficulty Tier Definitions}
\label{app:difficulty}

Each of the 33 question templates is assigned to one of three
difficulty tiers based on the computational steps required to
derive the answer from the backtest output.

\paragraph{Easy (9 templates).}
Direct single-row or single-scalar lookups from the backtest
dataframe, requiring no aggregation or multi-step reasoning.
Examples: first close price, volume or broker cash on a given
date, SMA value, position size, crossover/signal counts.

\paragraph{Medium (11 templates).}
Conditional row selection or simple aggregation over the
dataframe, requiring filtering by a condition or combining
multiple fields. Examples: first buy execution, profitable
vs.\ losing trade counts, net P/L, total commission, days in
position, peak portfolio date, P/L of a specific closed trade.

\paragraph{Hard (13 templates).}
Multi-step derived metrics requiring non-trivial computation
over the full time series or running a second backtest with
alternative parameters. Examples: maximum drawdown
(peak-to-trough), ROI, win rate, profit factor, annualized
return, Calmar ratio, exposure-adjusted return, drawdown
recovery days, golden cross counting within a sub-period,
and comparative questions across different SMA parameters.
\newpage
\section{Question Templates}
\label{app:templates}

Table~\ref{tab:templates} lists all 33 MCQ templates with their
difficulty tier and a brief description of what each question asks.

\begin{table}[h]
\caption{All 33 MCQ question templates.}
\label{tab:templates}
\centering
\small
\begin{tabular}{llp{10cm}}
\toprule
Tier & Template & Description \\
\midrule
E & first\_close & First close price \\
E & volume & Volume on a date \\
E & broker\_cash & Broker cash on a date \\
E & portfolio\_value & Portfolio value on a date \\
E & sma & SMA value on a date \\
E & max\_shares & Max affordable shares \\
E & position\_on\_date & Position size on a date \\
E & crossovers & Crossover event count \\
E & buy\_sell\_signals & Buy/sell signal counts \\
\midrule
M & first\_buy & First buy: shares, price, date \\
M & trade\_outcomes & Profitable/losing trade counts \\
M & net\_pnl & Final net P/L \\
M & commission & Commission + order count \\
M & total\_shares\_traded & Total shares traded \\
M & days\_in\_position & Days in/out of position \\
M & peak\_portfolio\_date & Peak portfolio date/value \\
M & best\_trade\_day & Best closed-trade day \\
M & strategy\_action\_on\_date & Buy/sell/hold on a date \\
M & nth\_trade\_pnl & Nth trade realized P/L \\
M & cash\_after\_nth\_order & Cash after nth order \\
\midrule
H & max\_drawdown & Max drawdown (\$) \\
H & roi\_percentage & ROI \% \\
H & win\_rate & Win rate \% \\
H & profit\_factor & Profit factor \\
H & annualized\_return & Annualized return \\
H & avg\_holding\_period & Avg holding period \\
H & golden\_cross\_count & Golden crosses in subperiod \\
H & max\_drawdown\_window & Drawdown window details \\
H & drawdown\_recovery\_days & Recovery days \\
H & calmar\_ratio & Calmar ratio \\
H & exposure\_adjusted\_return & Exposure-adj.\ return \\
H & sma\_comparison & Compare P/L across params \\
H & signal\_diff & Compare signals across params \\
\bottomrule
\end{tabular}
\end{table}

\newpage
\section{Example Questions by Difficulty}
\label{app:examples}

The following three examples are drawn from the 160-question base
pool (SMA Crossover strategy, AAPL, 2020-01-01 to 2024-01-01).
Each question is preceded by the standard configuration preamble
shown in Appendix~\ref{app:config}.

\paragraph{Easy: first close price.}\mbox{}

{\small\begin{verbatim}
What was the first close price observed
for AAPL in this backtest window?
A. $58.45   B. $95.86
C. $75.09   D. $96.60
Answer: C
\end{verbatim}}

\paragraph{Medium: first buy execution.}\mbox{}

{\small\begin{verbatim}
For the first buy execution, how many
shares were bought, at what closing
price, and on which date?
A. 668 shares at $71.67 on 2020-04-16
B. 478 shares at $177.15 on 2023-10-17
C. 443 shares at $186.40 on 2023-11-10
D. 396 shares at $178.96 on 2022-03-29
Answer: A
\end{verbatim}}

\paragraph{Hard: maximum drawdown.}\mbox{}

{\small\begin{verbatim}
What was the maximum portfolio drawdown
(peak-to-trough) during the backtest?
A. $28843.29   B. $35256.48
C. $34032.42   D. $23351.09
Answer: A
\end{verbatim}}

\section{Backtrader Plot}
\label{app:backtrader_plot}

\begin{figure}[h]
\centering
\includegraphics[width=0.8\textwidth]{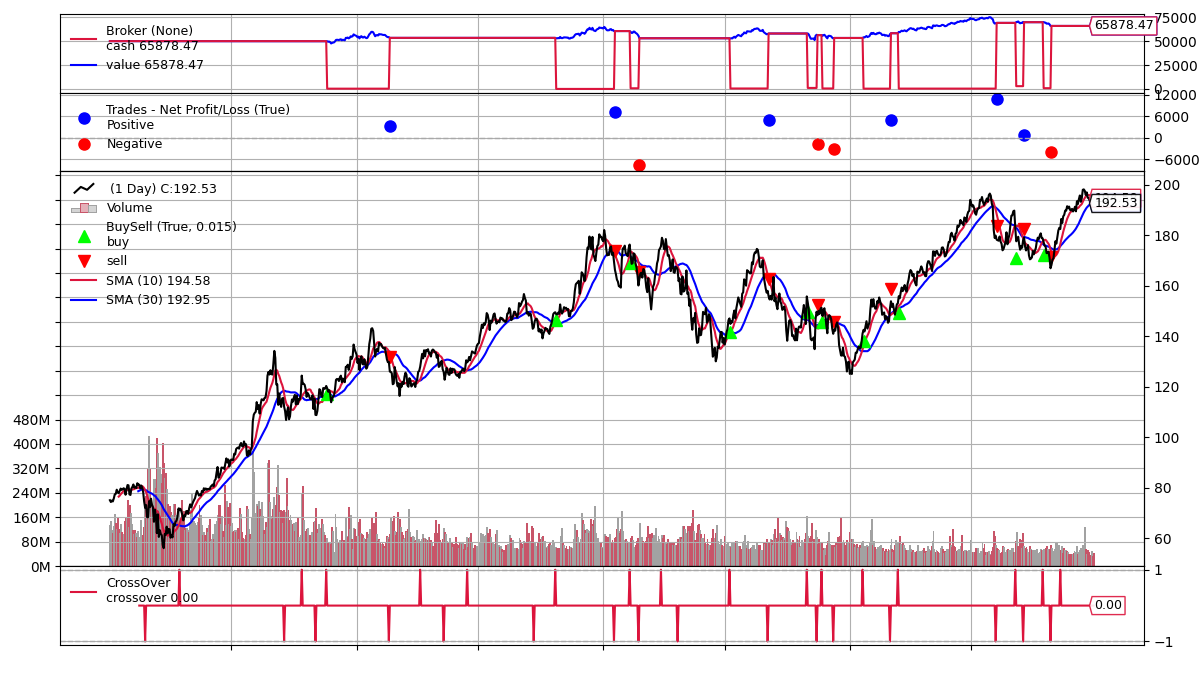}
\caption{Sample backtrader plot for SMA Crossover on AAPL
(2020 to 2024) using the default pipeline configuration. Buy and
sell markers, SMA indicators, trade profit, and portfolio value
are shown.}
\label{fig:backtrader_plot}
\end{figure}

Figure~\ref{fig:backtrader_plot} shows a sample backtrader plot
generated with the default pipeline configuration
(Appendix~\ref{app:config}): SMA Crossover strategy on AAPL from
2020-01-01 to 2024-01-01. The plot displays the price series with
buy/sell markers, the fast and slow SMA indicators, trade profit
markers, and the portfolio value over time. The visual structure of
this output directly inspired the MCQ template design: easy
questions ask about values readable from individual data points
(close price, volume, SMA value), medium questions target
trade-level events visible in the markers (first buy, trade P/L),
and hard questions require derived metrics computed over the full
series (drawdown, ROI, Sharpe ratio).

\newpage
\section{Pipeline Code}
\label{app:pipeline}

The pipeline is invoked through a single CLI entry point that
runs two stages sequentially: question generation followed by
ground-truth verification.

\paragraph{Build stage.}
The MCQ builder accepts a backtest configuration (symbol, date
range, initial cash, commission rate, stake size, strategy name,
and strategy parameters), runs a Cerebro backtest, extracts
summary statistics and daily records from the instrumented
strategy, and invokes the appropriate question factory for each
selected template. Each factory derives the correct answer from the
backtest output and generates three distractor options using
randomized multiplicative jitter. The builder writes each question
as a JSONL record containing the rendered question text with a
configuration preamble, four labeled options (A/B/C/D), the correct
answer letter, the question type, the difficulty tier, the strategy
metadata, and the full backtest configuration.

\paragraph{Check stage.}
An independent checker re-runs the same backtest from scratch using
only the configuration stored in the JSONL record, re-derives the
expected answer for each question type, and compares it against the
recorded correct option. Any mismatch flags a verification failure,
catching silent bugs such as template drift, rounding
inconsistencies, or changes to the underlying strategy logic.

{\small\begin{verbatim}
def run_pipeline():
    # STEP 1: Generate questions
    builder_main(config_argv)

    # STEP 2: Verify ground truth
    results = check_mcq_file(output_path)

    passed = sum(1 for r in results if r["match"])
    failed = len(results) - passed

    if failed > 0:
        # Flag mismatches for inspection
        for r in results:
            if not r["match"]:
                print(f"Q{r['index']}: "
                      f"stated={r['stated_answer']} "
                      f"computed={r['computed_answer']}")
        return 1
    return 0
\end{verbatim}}

The builder accepts configuration via CLI flags or a JSON file:

{\small\begin{verbatim}
python backtrader_MCQ_pipeline.py \
    --symbol AAPL \
    --start-date 2020-01-01 \
    --end-date 2024-01-01 \
    --initial-cash 50000 \
    --commission-rate 0.001 \
    --strategy sma_crossover \
    --num-questions 30 \
    --seed 42 \
    --difficulty all \
    -o output/AAPL_mcq_30.jsonl
\end{verbatim}}

An interactive mode (-i) walks the user through every
configuration field, and --all-templates generates one
question per template for base pool construction.

\newpage
\section{Mining Pipeline Code}
\label{app:mining_pipeline}

The generator-solver filtering pipeline
(Section~\ref{sec:filtering_pipeline}) is implemented as a
standalone CLI tool that orchestrates three components: a generator
agent, a verifier subprocess, and two solver agents (no-tools and
with-tools). The main loop runs a configurable number of mining
attempts (default 10), feeding the generator a history of prior
outcomes to discourage repeating easy patterns.

\paragraph{Mining loop.}
Each attempt proceeds through four stages: generation, verification,
MCQ conversion, and multi-stage filtering. The generator prompt
includes a running history of prior attempts and their decisions
(accepted, discarded as easy, or failed), enabling within-session
learning.

{\small\begin{verbatim}
for attempt in range(1, args.attempts + 1):
    prompt = build_generator_prompt(history)
    invo = invoke_agent(generator, prompt,
        cwd=generator_dir, timeout=args.generator_timeout,
        model=args.generator_model)

    spec = parse_generator_spec(invo.stdout)
    mined, base_run, perturb_runs, reasons = \
        build_mined_question(args, spec, attempt,
            attempt_dir, base_seed=attempt_seed)

    # Stage 1: no-tool solver screens for easy questions
    no_tools = run_no_tools(args, question_path, attempt_dir)
    if runner_correct(no_tools.get("record")):
        decision = "easy_discarded"
        continue

    # Stage 2: tool-augmented solver validates remainder
    with_tools = run_with_tools(args, question_path, attempt_dir)
    decision = "tool_solver_correct" \
        if runner_correct(with_tools.get("record")) \
        else "tool_solver_wrong"
    history.append(attempt_summary)
\end{verbatim}}

\paragraph{Generator output parsing.}
The generator returns three delimited blocks (question stem,
verification code, rationale). The parser extracts each block via
regex matching on the custom tag format:

{\small\begin{verbatim}
def parse_generator_spec(raw_text):
    text = logical_text(raw_text)
    question = extract_block(text, "QUESTION")
    code = extract_block(text, "CODE")
    rationale = extract_block(text, "RATIONALE") or ""
    if not question:
        raise ValueError("missing <<<QUESTION>>> block")
    if not code:
        raise ValueError("missing <<<CODE>>> block")
    return QuestionSpec(question, code, rationale)
\end{verbatim}}

\paragraph{Verifier execution.}
The verification code produced by the generator is written to a
temporary file and executed as a subprocess with the seed passed
via the MCQ\_SEED environment variable. A 120-second timeout
prevents runaway executions. The verifier output is parsed for
structured answer lines (QUESTION\_JSON and ANSWER\_VALUE\_JSON):

{\small\begin{verbatim}
def run_verifier(code, *, seed, work_dir, timeout=120,
                 python_path=None):
    script_path = work_dir / "verifier.py"
    script_path.write_text(code)
    env = os.environ.copy()
    env["MCQ_SEED"] = str(seed)
    proc = subprocess.run(
        [str(python_path), str(script_path)],
        capture_output=True, text=True, timeout=timeout,
        cwd=str(work_dir), env=env)
    question, answer_value, answer_letter = \
        parse_verifier_output(proc.stdout)
    return VerificationRun(seed, proc.stdout, proc.stderr,
        proc.returncode, timed_out=False,
        question=question, answer_value=answer_value,
        answer_letter=answer_letter)
\end{verbatim}}

\paragraph{MCQ conversion via seed perturbation.}
Distractors are generated by re-running the same verification code
with perturbed seed values (up to 8 perturbations, default step
size 1009). When fewer than four unique values emerge from
perturbation, numeric jitter ($\pm$10\%, $\pm$20\%) is applied to
the correct answer. Options are shuffled deterministically:

{\small\begin{verbatim}
def build_options_from_values(*, correct_value,
                              distractor_values, shuffle_seed):
    choices = []
    seen = set()
    add(correct_value)
    for v in distractor_values:
        add(v)
        if len(choices) >= 4: break

    # Fallback: numeric jitter if too few distinct values
    if len(choices) < 4 and is_numeric(correct_value):
        for delta in [0.1, -0.1, 0.2, -0.2]:
            add(correct_value * (1.0 + delta))
            if len(choices) >= 4: break

    rng = random.Random(shuffle_seed)
    rng.shuffle(choices[:4])
    options = {letter: choices[i]
               for i, letter in enumerate("ABCD")}
    answer = next(l for l, v in options.items()
                  if v == correct_value)
    return options, answer, choices
\end{verbatim}}

\paragraph{CLI invocation.}
The mining pipeline accepts configuration via CLI flags:

{\small\begin{verbatim}
python mine_backtrader_questions.py \
    --attempts 10 \
    --generator-agent cursor \
    --generator-model gpt-5.5-medium \
    --no-tools-model gpt-5.4 \
    --with-tools-model gpt-5.4 \
    --generator-timeout 600 \
    --code-timeout 120 \
    --no-tools-timeout 120 \
    --with-tools-timeout 180 \
    --choice-perturbations 8 \
    --seed-step 1009
\end{verbatim}}

All artifacts (generator output, verifier runs, solver reports,
and per-attempt summaries) are persisted as JSONL on disk, and a
final summary reports acceptance counts, token usage per role, and
per-model invocation counts.
\newpage
\paragraph{Persistence and directory layout.}
The pipeline persists all state as flat JSONL files organized into
three directories under a timestamped run identifier, avoiding
external database dependencies:

{\small\begin{verbatim}
db/
  mcq/       # question records with verification
  attempts/  # per-question solver attempts
  summary/   # per-attempt and final summaries
logs/
  mining_YYYYMMDD_HHMMSS/
    attempt_001/   # generator output, verifier
    attempt_002/   #   runs, solver reports
    ...
\end{verbatim}}

Each mining session receives a unique run identifier constructed
from the current timestamp and the number of rounds. All output
files for a run share the same prefix, making it straightforward
to correlate questions, attempts, and summaries:

{\small\begin{verbatim}
def make_run_id(*, attempts, run_name=None):
    timestamp = datetime.now().strftime("%Y%m%d_%H%M%S")
    base = f"mining_{timestamp}_rounds_{attempts}"
    if run_name:
        base = f"{base}_{safe_path_segment(run_name)}"
    if not _run_files_exist(base):
        return base
    suffix = 2
    while True:
        candidate = f"{base}_{suffix}"
        if not _run_files_exist(candidate):
            return candidate
        suffix += 1
\end{verbatim}}

\paragraph{Content-based deduplication.}
Each question is assigned a deterministic identifier by hashing
its normalized text and four options with SHA-256. Whitespace
is collapsed before hashing so that formatting differences do
not produce distinct identifiers. When a question with an
existing hash is generated in a later round, it is skipped
automatically:

{\small\begin{verbatim}
def question_hash(question, options):
    parts = [_normalize_ws(question)]
    for letter in ("A", "B", "C", "D"):
        parts.append(
            f"{letter}:{_normalize_ws(options.get(letter, ''))}")
    return hashlib.sha256(
        "\n".join(parts).encode("utf-8")).hexdigest()
\end{verbatim}}

\paragraph{Well-posedness validation.}
Before entering the filtering stages, each candidate must pass a
well-posedness check that rejects malformed questions early:

{\small\begin{verbatim}
def evaluate_well_defined(*, question, options, code,
        verifier_returncode, verifier_timed_out, answer):
    reasons = []
    if not question or not question.strip():
        reasons.append("missing question text")
    for letter in ("A", "B", "C", "D"):
        if not options.get(letter, "").strip():
            reasons.append(f"missing option {letter}")
    if not code or not code.strip():
        reasons.append("missing verification code")
    if verifier_timed_out:
        reasons.append("verification code timed out")
    elif verifier_returncode != 0:
        reasons.append(f"non-zero exit {verifier_returncode}")
    if answer not in ("A", "B", "C", "D"):
        reasons.append("no single ground-truth answer")
    return len(reasons) == 0, reasons
\end{verbatim}}

\paragraph{Record schema.}
Each question record stored in the JSONL file contains: a
content-based hash identifier, the question text and four options,
the ground-truth answer letter, the full verification code and its
execution output, a well-posedness flag with failure reasons, the
pipeline decision (accepted, discarded as easy, or flagged for
human review), and a set of labels for downstream filtering. Solver
attempt records link back to the question identifier and store the
agent's answer, correctness, token usage, and evaluation mode
(no-tools screening or with-tools validation).

\newpage
\section{Pipeline Configuration}
\label{app:config}

The following shows the default configuration dictionary used by
the pipeline. Every field can be overridden via CLI flags, a JSON
config file, or the interactive mode.

{\small\begin{verbatim}
PIPELINE_CONFIG = {
    "symbol":          "AAPL",
    "start_date":      "2020-01-01",
    "end_date":        "2024-01-01",
    "initial_cash":    50000,
    "commission_rate": 0.001,
    "stake":           1000,
    "strategy":        "sma_crossover",
    "strategy_params": {"pfast": 10, "pslow": 30,
                        "target_percent": 1.0},
    "num_questions":   10,
    "seed":            42,
    "difficulty":      "easy",
    "output":          "backtrader_mcq.jsonl",
    "benchmark":       "backtrader",
    "author":          "MCQ_test",
}
\end{verbatim}}

\paragraph{Field descriptions.}
\begin{itemize}
\item \textbf{symbol}: Ticker symbol for the equity to backtest
  (e.g., AAPL, MSFT, GOOG).
\item \textbf{start\_date}, \textbf{end\_date}: Date range for the
  backtest window in YYYY-MM-DD format.
\item \textbf{initial\_cash}: Starting capital for the broker.
\item \textbf{commission\_rate}: Broker commission as a decimal
  fraction (e.g., 0.001 = 0.1\%).
\item \textbf{stake}: Fixed order size in shares per trade.
\item \textbf{strategy}: Name of the trading strategy. One of: SMA Crossover, Rolling Window Mean, Exponential Weighted Moving Average, RSI Strategy, MACD Crossover.
\item \textbf{strategy\_params}: Strategy-specific parameters as a
  dictionary (e.g., SMA periods, RSI thresholds).
\item \textbf{num\_questions}: Number of MCQ questions to generate.
\item \textbf{seed}: Random seed for reproducibility.
\item \textbf{difficulty}: Difficulty pool to draw from: ``easy'',
  ``medium'', ``hard'', ``all'' (balanced), or a comma-separated
  combination like ``easy,medium''.
\end{itemize}

\newpage
\section{Sampling and Reproducibility Details}
\label{app:sampling}

The single random seed passed to the pipeline controls five
stochastic aspects of question generation:
\begin{enumerate}
\item Selecting which question templates appear in the output.
\item Generating distractor answers via multiplicative jitter
  around the correct value.
\item Shuffling the A/B/C/D option ordering.
\item Picking sample dates for date-parameterized questions.
\item Selecting alternative parameter sets for comparison questions.
\end{enumerate}

When multiple difficulty levels are requested, the total question
count is split as evenly as possible across the selected levels.
Within each level, templates are sampled without replacement when
the quota is smaller than the pool; otherwise the pool is cycled
with replacement to fill the remaining slots. This balanced
allocation ensures that no single difficulty tier dominates the
output.

An ``all-templates'' mode generates exactly one question per
template for the selected difficulty levels, producing a base pool
that covers every question type. The 160-question dataset released
with this paper is such a base pool, generated under the default
configuration (Appendix~\ref{app:config}).

Because the stock market is highly stochastic, different symbols
and date ranges produce entirely different price data, trade
sequences, and ground-truth answers, yielding fresh question sets
suitable for independent evaluation runs.

\section{Prompt Templates}
\label{app:prompts}

\paragraph{With-tools solver prompt.}
The following prompt wraps each MCQ when run with code-execution
tools enabled.

{\footnotesize\begin{verbatim}
You are answering questions about a trading
strategy and stock market behavior. Use the
Backtrader package to code, compute or verify all
answers.

Configuration for this question group:
- commission_rate: 0.001
- end_date: 2024-01-01
- initial_cash: 50000.0
...

Use the config and strategy setting for this
question. Run the Backtrader strategy when
positions, orders, trades, drawdowns, or alternate
parameters matter; all answers should come from the
resulting backtest dataframe values and the formula
stated in the question.

[question text and options]

Respond in the form <<< X >>> where X is A, B, C,
or D.
\end{verbatim}}

\paragraph{No-tools prompt.}
The agent is explicitly prohibited from using any external
capability and must answer from internal reasoning alone.

{\footnotesize\begin{verbatim}
You are a no-tools multiple-choice answerer. This
is a one-shot API chat completion, not an agent
task.
Do not use tools. Do not call web search, web
fetch, Bash, Python, MCPs, file search, file read,
file write, browser, workspace inspection, or any
other external capability. Use only the question
text and your internal reasoning.
Reply with exactly <<< X >>> where X is one of A,
B, C, or D. No explanation, no code, no extra
text.

[question text and options]
\end{verbatim}}

\newpage
\paragraph{Generator prompt.}
\label{app:generator_prompt}
The following prompt is sent to the generator LLM for each
mining round. The miner builds the final four MCQ choices from
the verifier output; the generator provides only the question
stem and standalone verification code.
The \{HISTORY\} placeholder is filled at runtime with prior
mining outcomes to discourage repeating easy patterns.

{\footnotesize\begin{verbatim}
You are generating one Backtrader MCQ spec.
The miner builds the final four choices. Provide:
1. A question stem (no A/B/C/D options).
2. Standalone Python verifier code computing the answer.

The verifier runs with different MCQ_SEED values. Use
the seed to perturb inputs (ticker, dates, strategy
params, cash, commission, metric). The base seed
defines the true question/answer; perturbed runs yield
distractor values.

Verifier output protocol:
- Print one QUESTION_JSON: ... line (JSON string).
- Print one ANSWER_VALUE_JSON: ... line (JSON string).
- Do not print A/B/C/D or ANSWER: X.

Requirements:
- About backtrader strategies or computed backtest
  metrics; must require code, not general finance facts.
- Use backtrader, pandas, numpy, optionally yfinance.
- Prefer deterministic data or generated OHLCV.
- Verifier must be standalone (no project imports).
- Keep base and perturbed runs in the same question
  family. Answer: compact scalar ($1234.56, 17, etc.).

Prior mining outcomes: {HISTORY}

Output exactly these blocks, no markdown fences:

<<<QUESTION>>>
<question stem, no options>
<<</QUESTION>>>

<<<CODE>>>
import json, os
seed = int(os.environ.get("MCQ_SEED", "0"))
# Compute scenario and answer here.
print("QUESTION_JSON:", json.dumps(question))
print("ANSWER_VALUE_JSON:", json.dumps(answer_value))
<<</CODE>>>

<<<RATIONALE>>>
Why this requires executable Backtrader reasoning.
<<</RATIONALE>>>
\end{verbatim}}

\newpage
\section{Agent Invocation Details}
\label{app:invocation}

\paragraph{Cursor Agent CLI.}
Each MCQ is dispatched to Cursor Agent via a non-interactive CLI
invocation of the form:

{\small\begin{verbatim}
agent -p --output-format json --yolo \
      --trust --workspace <session-dir> \
      [--model <model>] "<prompt>"
\end{verbatim}}

The \emph{--yolo} flag auto-approves tool use without pausing
for interactive confirmation. Tool availability is determined by
Cursor Agent's standard tool set, which includes shell execution,
file reading and writing, codebase search, and web access.
The \emph{--workspace} flag isolates each question to its own
session directory, enforcing the i.i.d.\ condition.

The core invocation function:

{\small\begin{verbatim}
def run_cursor_agent(agent_bin, prompt,
                     timeout, cwd, extra):
    cmd = [
        agent_bin, "-p",
        "--output-format", "json",
        "--yolo", "--trust",
        "--workspace", str(cwd),
        *extra,
        prompt,
    ]
    return subprocess.run(
        cmd, capture_output=True, text=True,
        timeout=timeout, check=False, cwd=cwd,
        stdin=subprocess.DEVNULL,
    )
\end{verbatim}}

\paragraph{No-tools enforcement.}
Cursor does not expose a documented ``disable all tools'' switch.
To enforce a no-tools baseline, we use the Cursor SDK
(@cursor/sdk) with three layered restrictions:
(i) the agent is created with an empty MCP server map, no
sub-agents, and sandboxing enabled;
(ii) the prompt explicitly prohibits all external tools;
(iii) the runner inspects every streaming event and, when it
detects a tool-call-shaped SDK event, logs the event to a
tool events file, sets the question status to ``tool attempted,''
and cancels the run if cancellation is supported. These
detections represent attempted or blocked tool calls, not
successful tool use. Affected questions are not answered
normally, which is why the answered percentage drops below
100\% for some difficulty levels in the no-tools results.
Questions flagged as ``tool attempted'' are excluded from
the no-tools accuracy calculation.

SDK agent initialization:

{\small\begin{verbatim}
const agent = await Agent.create({
  apiKey: process.env.CURSOR_API_KEY,
  model: { id: args.model },
  local: {
    cwd: process.cwd(),
    settingSources: [],
    sandboxOptions: { enabled: true },
  },
  mcpServers: {},
  agents: {},
});
\end{verbatim}}

\paragraph{Token estimation.}
When exact usage metadata is available in streaming events or the
final result, the runner extracts it (supporting both snake\_case
and camelCase field names). When exact counts are unavailable, the
runner falls back to a regex-based approximation that tokenizes
the prompt and response into word and punctuation chunks and
scales by 1.25x. Estimated values are tagged as ``estimated''
in the report.

\newpage
\section{Agent Backend Registry}
\label{app:agent_backends}

The adapter layer provides a uniform interface for invoking
different coding-agent CLIs. Each backend is registered as an
AgentDef dataclass that specifies the binary name, prompt
filename, and a run function. The pipeline selects a backend by
name and calls it through the same interface regardless of the
underlying CLI conventions.

\paragraph{Registry design.}
Each backend is an AgentDef (name, binary path, prompt filename,
run function, optional extra flags). A global registry maps
backend names to definitions; adding a new backend requires only
implementing a run function with a fixed signature (binary path,
prompt string, timeout, working directory, extra flags) and
registering it.

\paragraph{Supported backends.}
The registry currently includes four backends:
\emph{Cursor} (Agent CLI with JSON output, auto-approved tool use,
per-question workspace isolation);
\emph{Claude Code} (Bash tool access with permission skipping for
batch evaluation);
\emph{OpenAI Codex} (exec subcommand with JSON output);
\emph{GitHub Copilot} (automatic CLI variant detection across
different flag conventions, with result caching).
All experiments in this paper use the Cursor backend.

\paragraph{Token usage extraction.}
Each CLI produces JSON in a different format. A chain of
format-specific adapters (Claude, Cursor, Codex, OpenAI)
normalizes token counts into a common schema (input tokens,
output tokens, cache read tokens, model). The extractor tries
each adapter in sequence until one matches, enabling uniform
cost tracking across heterogeneous backends.

\section{MCQ Evaluation Protocol}
\label{app:eval_protocol}

Due to token cost and computational constraints, we evaluate on a
balanced subset of 30 questions (10 per difficulty tier) rather
than the full 160-question pool. The subset covers all five
strategies. Each question is presented in a fresh sandbox: the
agent starts from scratch with only the MCQ prompt and has no
access to prior questions or answers, enforcing the i.i.d.\
condition across the evaluation set.

Each agent is run in two modes: \emph{with tools} (code execution
enabled via shell and package installation) and
\emph{without tools} (API-only, no external capabilities). For each
run we record: (i) accuracy at each difficulty level and overall,
(ii) whether the agent wrote and executed code (to verify the
tool-use mode operated correctly), (iii) elapsed time per
question, and (iv) input and output token counts.

\paragraph{Multi-run averaging (no-tools, curated benchmark).}
Because LLM inference is inherently stochastic (sampling
temperature, non-deterministic decoding), a single evaluation pass
can be misleading. For the curated benchmark, all 11 no-tools
models are run 10 times each; reported accuracies are 10-run
averages with standard deviations. For example, Opus 4.6 scored
76.7\% in a single run but averages 64.3\% over 10 runs, an
overestimate of more than 12 percentage points. Multi-run averaging
is especially important for no-tools evaluation, where the model
relies entirely on internal reasoning with no execution grounding.

\paragraph{Single-pass evaluation (with-tools and mined MCQs).}
With-tools runs require 83 to 103 minutes per 30-question set
because the agent iteratively writes, executes, and debugs code.
This high cost limits with-tools evaluation on the curated
benchmark to a single pass per model configuration. Similarly,
no-tools evaluation on the 38 mined MCQs is a single run per model
due to the exploratory nature of the mining evaluation. Single-run
variance on 38 questions can be substantial: Opus 4.7 scored 68.4\%
in one run and 52.6\% in another on the same mined set.

\newpage
\section{Mined Question Domains}
\label{app:mcq-domains}

The generator/solver pipeline produces questions spanning several
distinct domains. We classify the 98 well-posed mined questions
into eight categories based on the trading mechanics each question
exercises. Below we describe each type in detail.

\paragraph{Plain MA crossover to final equity (43 questions, 47\% accepted).}
These ask for the final broker value after a deterministic
SMA/EMA crossover strategy. The core mechanics are indicator
warm-up, crossover timing, fixed-size market orders, commissions,
and final mark-to-market value. This is the most common generated
type; many are discarded because the no-tools solver can
sometimes infer the correct option from a familiar template or
answer-choice structure.
\emph{Financial relevance:} MA crossover is the canonical
trend-following signal used in production systematic trading;
correctly computing final equity after commissions is the most
basic validation any backtesting workflow must pass.

\paragraph{MA crossover + timed/percent exits (19 questions, 21\% accepted).}
These add exit rules such as ``sell after holding $N$ bars'' or
``sell after a fixed-percent gain/loss.'' Despite appearing more
complex, the added rule is usually explicit and local; in this
sample they skew easy, suggesting that simple exit overlays do
not reliably force tool use unless they interact with pending
orders or subtle Backtrader state.
\emph{Financial relevance:} timed and percent-based exit rules
are standard risk management overlays in production strategies;
they interact with position sizing and can change P\&L
substantially when applied to volatile instruments.

\paragraph{Bracket / trailing-stop order mechanics (15 questions, 60\% accepted).}
These involve Backtrader-specific order behaviour: parent/child
bracket orders, stop/limit children, StopTrail, moving
stop levels, cancellation logic, or interactions between
protective orders and crossover exits. This is the strongest
category; questions are accepted 60\% of the time because they
require precise knowledge of Backtrader's event loop, order
notifications, execution timing, and state transitions.
\emph{Financial relevance:} bracket orders and trailing stops
are the primary downside-protection mechanisms in live trading.
Misunderstanding their fill semantics (e.g., which child order
triggers first, how cancellation propagates) can lead to
unhedged positions or unexpected margin calls.

\paragraph{Final equity with slippage (6 questions, 17\% accepted).}
Final-value backtests where slippage is part of the broker
setup. Slippage alone did not make questions difficult enough in
this sample, likely because its numerical effect is small or the
adjusted value is obvious from the answer choices.
\emph{Financial relevance:} slippage modelling is critical for
realistic backtest-to-live performance translation;
underestimating slippage is one of the most common causes of
strategy degradation in production.

\paragraph{MA/ATR regime with ATR-based exits (5 questions, 40\% accepted).}
Moving-average entry logic with stops or take-profits derived
from ATR values. These require computing a dynamic volatility
indicator and applying it to exit logic; they are moderately
promising but the sample is small.
\emph{Financial relevance:} ATR-scaled exits adapt to market
volatility regimes, a technique widely used in managed futures
and CTA strategies.

\paragraph{Analyzer / non-equity metric (5 questions, 20\% accepted).}
Questions asking for max drawdown, net profit, closed trade
count, or another analyzer-style statistic rather than final
portfolio value. Often discarded as easy because some metrics are
near-zero, integer-like, or guessable from the scenario.
\emph{Financial relevance:} drawdown and trade-level statistics
are among the most scrutinized metrics in fund due diligence;
an agent that misreports max drawdown could lead to incorrect
risk assessments.

\paragraph{Inline OHLC / CSV feed (3 questions, 33\% accepted).}
A small OHLCV table is pasted directly into the prompt. Making
data visible can help readability but may also make the problem
easier for a no-tools solver when the table is short.

\paragraph{Breakout with filters (2 questions, 0\% accepted).}
Entry rules like buying when close breaks above the previous
$N$-bar high, often with indicator filters. Both examples were
discarded as easy; the sample is too small to draw conclusions.
\emph{Financial relevance:} breakout strategies are a staple
of momentum-based trading and represent an important domain for
future mining expansion.

\end{document}